\documentclass{article}

     \usepackage[final]{neurips_2020}

\usepackage[utf8]{inputenc} 
\usepackage[T1]{fontenc}    
\usepackage{hyperref}       
\usepackage{url}            
\usepackage{booktabs}       
\usepackage{amsfonts}       
\usepackage{nicefrac}       
\usepackage{microtype}      

\usepackage{amsthm,mathtools}

\theoremstyle{definition}

\usepackage{xcolor}

\setcitestyle{round}

\title{SuperSuit: Simple Microwrappers for Reinforcement Learning Environments}

\author{%
  Justin K. Terry\\
  Department of Computer Science\\
  University of Maryland, College Park\\
  College Park, MD 20742 \\
  \texttt{jkterry@umd.edu} \\
  \And
  Benjamin Black\\
  Department of Computer Science\\
  University of Maryland, College Park\\
  College Park, MD 20742 \\
  \texttt{bblack1@umd.edu} \\
  \And
  Ananth Hari \\
  Department of Computer Engineering\\
  University of Maryland, College Park\\
  College Park, MD 20742 \\
  \texttt{ahari1@umd.edu} \\
}

\begin{document}

\maketitle

\begin{abstract}
  In reinforcement learning, wrappers are universally used to transform the information that passes between a model and an environment. Despite their ubiquity, no library exists with reasonable implementations of all popular preprocessing methods. This leads to unnecessary bugs, code inefficiencies, and wasted developer time. Accordingly we introduce SuperSuit, a Python library that includes all popular wrappers, and wrappers that can easily apply lambda functions to the observations/actions/reward. It's compatible with the standard Gym environment specification, as well as the PettingZoo specification for multi-agent environments. The library is available at \url{https://github.com/PettingZoo-Team/SuperSuit}, and can be installed via pip.
\end{abstract}

\section{Introduction}
Applying transformations to information passing between a model and an environment in reinforcement learning is an integral part of every major experimental work in the field \citep{mnih2013playing, vinyals2019grandmaster, silver2017mastering, berner2019dota}. Techniques popular on Atari environments include scaling down observations with image processing methods or making the observation greyscale to reduce processing time with neural networks, ``stacking'' frames together to help establish velocity, or skipping frames to increase training speed \citep{mnih2013playing}.

These ``wrappers'' are very useful, but using them in practice has pain points. For code modularity, ease of debugging, and ease of hyper-parameter tuning, it's generally preferable to define the wrapper function(s) outside the environment. Ideally these very commonly used functions would be distributed in a library, so that the implementation used is as fast as possible. This is fairly important considering how many times it would be called in large research projects.

Gym \citep{Gym} has become the standard API and set of benchmark environments for single-agent reinforcement learning. PettingZoo \citep{pettingZoo2020} has recently been released, achieving similar goals for multi-agent reinforcement learning environments. The only existing library with wrappers for reinforcement learning are those included inside Gym, but those are primarily the initially popular wrappers for Atari preprocessing \citep{mnih2013playing}. Newer preprocessing methods for Atari \citep{machado18arcade}, other types of environments, or multi-agent environments are omitted. Many Gym wrappers are also missing ``quality of life'' features, like outputting arrays in a shape compatible with CNNs by default. Accordingly, people typically write their own wrappers themselves. This leads to lower code quality and performance throughout the field for such key functions, and makes the possibility of bugs greater. Accordingly, we've released the SuperSuit Python library to include all widely used wrappers for both Gym and PettingZoo environments. Each wrapper is a function that takes an environment object and returns one, and for clarity and modularity, only includes a single function, hence our terming them ``microwrappers''.

\section{Wrapper Methods}

The observation wrappers we include are:

\begin{itemize}
    \item Agent Indication (Multi-Agent Only) \citep{gupta2017cooperative}
    \item Color Reduction (Greyscaling, etc.)
    \item Flatten Observation
    \item Frame Skipping \citep{mnih2013playing}
    \item Frame Stacking \citep{mnih2013playing}
    \item Observation Delay
    \item Observation Normalization
    \item Observation Padding (Multi-Agent Only) \citep{TerryParameterSharing}
    \item Recast Observation Type
    \item Reshape Observation
    \item Resize 2D/3D Observation
\end{itemize}

The action wrappers we include are:

\begin{itemize}
    \item Action Clipping \citep{fujita2018clipped}
    \item Action Space Padding (Multi-Agent Only) \citep{TerryParameterSharing}
    \item Sticky Actions \citep{machado18arcade}
\end{itemize}

The only reward wrapper we include is:
\begin{itemize}
    \item Reward Clipping \citep{mnih2013playing}
\end{itemize}

Additionally, we introduce \emph{lambda wrappers} that take an environment and a lambda function as an argument and the lambda function to the environment it, allowing people to easily create custom wrappers. Separate lambda wrappers exist to apply functions to actions, observations, or rewards.

\section{Conclusion}
We introduce \emph{SuperSuit}, a Python library that includes reasonable implementations of all popular RL wrappers, for environments of both the Gym and PettingZoo API specification. This will allow researchers to conduct more computationally efficient experiments, to try new RL wrappers much more easily, and to reduce the likelihood of bugs due to one-off implementations. The library is available at \url{https://github.com/PettingZoo-Team/SuperSuit}, and can be installed via pip.

\begin{ack}
Justin Terry was supported in part by the QinetiQ Fundamental Machine Learning Fellowship.
\end{ack}

\bibliography{main}

\begin{thebibliography}{10}
\providecommand{\natexlab}[1]{#1}
\providecommand{\url}[1]{\texttt{#1}}
\expandafter\ifx\csname urlstyle\endcsname\relax
  \providecommand{\doi}[1]{doi: #1}\else
  \providecommand{\doi}{doi: \begingroup \urlstyle{rm}\Url}\fi

\bibitem[Berner et~al.(2019)Berner, Brockman, Chan, Cheung, D{\k{e}}biak,
  Dennison, Farhi, Fischer, Hashme, Hesse, et~al.]{berner2019dota}
Christopher Berner, Greg Brockman, Brooke Chan, Vicki Cheung, Przemys{\l}aw
  D{\k{e}}biak, Christy Dennison, David Farhi, Quirin Fischer, Shariq Hashme,
  Chris Hesse, et~al.
\newblock Dota 2 with large scale deep reinforcement learning.
\newblock \emph{arXiv preprint arXiv:1912.06680}, 2019.

\bibitem[Brockman et~al.(2016)Brockman, Cheung, Pettersson, Schneider,
  Schulman, Tang, and Zaremba]{Gym}
Greg Brockman, Vicki Cheung, Ludwig Pettersson, Jonas Schneider, John Schulman,
  Jie Tang, and Wojciech Zaremba.
\newblock Openai gym, 2016.

\bibitem[Fujita and Maeda(2018)]{fujita2018clipped}
Yasuhiro Fujita and Shin-ichi Maeda.
\newblock Clipped action policy gradient.
\newblock \emph{arXiv preprint arXiv:1802.07564}, 2018.

\bibitem[Gupta et~al.(2017)Gupta, Egorov, and
  Kochenderfer]{gupta2017cooperative}
Jayesh~K Gupta, Maxim Egorov, and Mykel Kochenderfer.
\newblock Cooperative multi-agent control using deep reinforcement learning.
\newblock In \emph{International Conference on Autonomous Agents and Multiagent
  Systems}, pages 66--83. Springer, 2017.

\bibitem[Machado et~al.(2018)Machado, Bellemare, Talvitie, Veness, Hausknecht,
  and Bowling]{machado18arcade}
Marlos~C. Machado, Marc~G. Bellemare, Erik Talvitie, Joel Veness, Matthew~J.
  Hausknecht, and Michael Bowling.
\newblock Revisiting the arcade learning environment: Evaluation protocols and
  open problems for general agents.
\newblock \emph{Journal of Artificial Intelligence Research}, 61:\penalty0
  523--562, 2018.

\bibitem[Mnih et~al.(2013)Mnih, Kavukcuoglu, Silver, Graves, Antonoglou,
  Wierstra, and Riedmiller]{mnih2013playing}
Volodymyr Mnih, Koray Kavukcuoglu, David Silver, Alex Graves, Ioannis
  Antonoglou, Daan Wierstra, and Martin Riedmiller.
\newblock Playing atari with deep reinforcement learning.
\newblock \emph{arXiv preprint arXiv:1312.5602}, 2013.

\bibitem[Silver et~al.(2017)Silver, Schrittwieser, Simonyan, Antonoglou, Huang,
  Guez, Hubert, Baker, Lai, Bolton, et~al.]{silver2017mastering}
David Silver, Julian Schrittwieser, Karen Simonyan, Ioannis Antonoglou, Aja
  Huang, Arthur Guez, Thomas Hubert, Lucas Baker, Matthew Lai, Adrian Bolton,
  et~al.
\newblock Mastering the game of go without human knowledge.
\newblock \emph{nature}, 550\penalty0 (7676):\penalty0 354--359, 2017.

\bibitem[Terry et~al.(2020{\natexlab{a}})Terry, Black, Jayakumar, Hari, Santos,
  Dieffendahl, Williams, Ravi, Lokesh, Horsch, and Patel]{pettingZoo2020}
Justin~K Terry, Benjamin Black, Mario Jayakumar, Ananth Hari, Luis Santos,
  Clemens Dieffendahl, Niall Williams, Praveen Ravi, Yashas Lokesh, Caroline
  Horsch, and Dipam Patel.
\newblock Petting{Z}oo.
\newblock \url{https://github.com/PettingZoo-Team/PettingZoo},
  2020{\natexlab{a}}.
\newblock GitHub repository.

\bibitem[Terry et~al.(2020{\natexlab{b}})Terry, Grammel, Hari, and
  Santos]{TerryParameterSharing}
Justin~K Terry, Nathaniel Grammel, Ananth Hari, and Luis Santos.
\newblock Parameter sharing is surprisingly useful for multi-agent deep
  reinforcement learning.
\newblock \emph{arXiv preprint arXiv:2005.13625}, 2020{\natexlab{b}}.

\bibitem[Vinyals et~al.(2019)Vinyals, Babuschkin, Czarnecki, Mathieu, Dudzik,
  Chung, Choi, Powell, Ewalds, Georgiev, et~al.]{vinyals2019grandmaster}
Oriol Vinyals, Igor Babuschkin, Wojciech~M Czarnecki, Micha{\"e}l Mathieu,
  Andrew Dudzik, Junyoung Chung, David~H Choi, Richard Powell, Timo Ewalds,
  Petko Georgiev, et~al.
\newblock Grandmaster level in starcraft ii using multi-agent reinforcement
  learning.
\newblock \emph{Nature}, 575\penalty0 (7782):\penalty0 350--354, 2019.

\end{thebibliography}

\end{document}